\documentclass{article}

\usepackage{svg} 
\usepackage{graphicx} 
\usepackage{algorithm}
\usepackage{algpseudocode}
\usepackage{tcolorbox} 
\usepackage{threeparttable} 
\usepackage{caption}        
\usepackage{booktabs}       
\usepackage{listings} 
\usepackage{tabularx}
\usepackage{tabularray}
\usepackage{changepage}
\usepackage{bbding}
\usepackage{diagbox}
\usepackage{setspace}
\usepackage[page,header]{appendix}
\usepackage{titling}
\newcommand{\LeftCommentOne}[1]{\Statex \hspace{\algorithmicindent} {\# #1}}

\algrenewcommand\algorithmicfor{\textbf{for}}
\algtext*{EndFor} 

\algrenewcommand\algorithmicif{\textbf{if}}
\algtext*{EndIf} 
\algrenewtext{algorithmicthen}{} 

\algtext*{EndProcedure} 

\usepackage[numbers,sort&compress]{natbib}
\usepackage[hidelinks]{hyperref}
\usepackage{longtable}
\usepackage{lscape}
\usepackage{amsmath,amsfonts}
\usepackage{cleveref}

\usepackage{authblk} 
\makeatletter
\renewcommand{\AB@affilsepx}{, }         
\makeatother

\usepackage{xcolor}
\definecolor{blue}{rgb}{0,0, 0.8}

\usepackage{geometry}
\title{MotifBench: A standardized protein design benchmark for motif-scaffolding problems}
\author[1]{Zhuoqi Zheng}
\author[1,2]{Bo Zhang}
\author[3,4]{Kieran Didi}
\author[5]{Kevin K. Yang}
\author[6]{Jason Yim}
\author[7]{Joseph L. Watson}
\author[1]{Hai-Feng Chen}
\author[8,9]{Brian L. Trippe\thanks{Correspondence to: btrippe@stanford.edu}}

\affil[1]{Department of Bioinformatics and Biostatistics, Shanghai Jiao Tong University}
\affil[2]{School of Life Sciences, Tsinghua University}
\affil[3]{Department of Computer Science, Oxford University}
\affil[4]{NVIDIA}
\affil[5]{Microsoft Research}
\affil[6]{Department of Computer Science, Massachusetts Institute of Technology}
\affil[7]{Department of Biochemistry, University of Washington}
\affil[8]{Department of Statistics, Stanford University}
\affil[9]{Stanford Data Science, Stanford University}

\begin{document}
\maketitle
\vspace{-0.5cm}

\begin{abstract}
The motif-scaffolding problem is a central task in computational protein design:
Given the coordinates of atoms in a geometry chosen to confer a desired biochemical function (a motif), the task is to identify diverse protein structures (scaffolds) that include the motif and maintain its geometry.
Significant recent progress on motif-scaffolding has been made due to computational evaluation with reliable protein structure prediction and fixed-backbone sequence design methods \citep{tischer2020design,wang2022scaffolding,trippe2022diffusion,watson2022broadly,ingraham2023illuminating,young2024diffusion,song2023joint,wu2024practical,chen2023amalga,didi2023framework,alamdari2023protein,zhang2023protein,hayes2024simulating,yim2024improved,lin2024out,wang2024dplm,frank2024scalable}.
However, significant variability in evaluation strategies across publications has hindered comparability of results, challenged reproducibility, and impeded robust progress.
In response we introduce MotifBench, comprising
    (1) a precisely specified pipeline and evaluation metrics, 
    (2) a collection of 30 benchmark problems, and 
    (3) an implementation of this benchmark and leaderboard at \url{github.com/blt2114/MotifBench}.
The MotifBench test cases are more difficult compared to earlier benchmarks (e.g.\ \citep{watson2022broadly}),
and include protein design problems for which solutions are known but on which, to the best of our knowledge, state-of-the-art methods fail to identify any solution.

\end{abstract}
\begin{figure}[ht]
    \centering
    \includegraphics[width=1.0\textwidth]{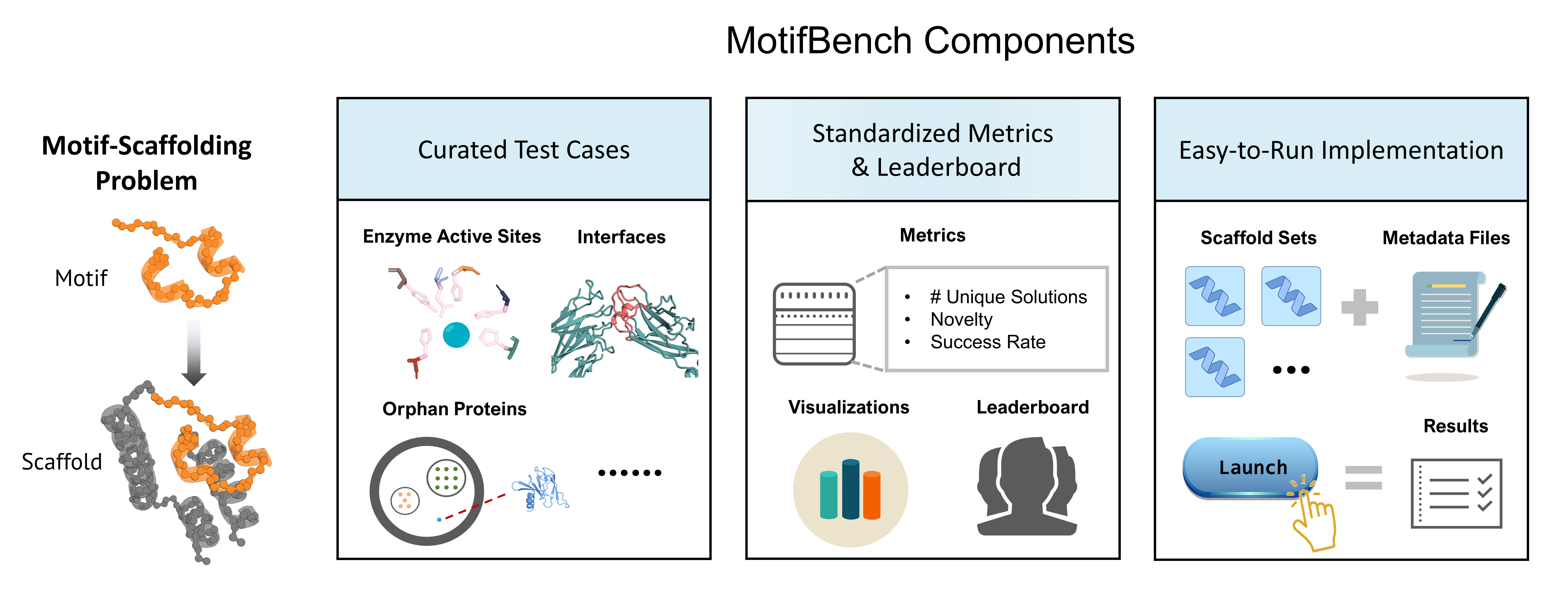} 
    \label{fig:overview} 
\end{figure}

\tableofcontents
\addtocontents{toc}{\protect\setcounter{tocdepth}{1}} 
\thispagestyle{empty}  
\clearpage             
\setcounter{page}{1}   
\pagebreak
\section{The motif-scaffolding problem: task specification and evaluation metrics}\label{sec:task_specification}

A motif-scaffolding method
takes a motif as input and returns a set of putatively compatible scaffolds as output.
This section details how motifs and scaffolds in MotifBench are specified,
proposes metrics by which a scaffold set is evaluated,
and describes how these metrics are computed.
\Cref{sec:considerations} describes considerations upon which these specifications and metrics were chosen.

\paragraph{Motif specification (inputs):}
A \emph{motif} is specified by the coordinates of the backbone atoms of several residues and (in some cases) the amino acid types of a subset of those residues.
\begin{itemize}
    \item {Motif atom coordinates are extracted from experimental structures deposited in the Protein Data Bank (PDB) \citep{berman2002protein}.}
    \item {Each problem comprises the N, C$\alpha$, and C backbone coordinates of several residues.}
    \item {The residues may be part of a single sequence-contiguous segment, or they may be divided across multiple segments.}
    \item {When the amino acid types at particular positions are thought to be important to the biochemical function of the motif, these types and positions are specified and may not be modified in designed scaffolds.  Side-chain heavy atom coordinates for these positions are also provided as optional inputs.}
\end{itemize}

\paragraph{Scaffold set specification (outputs):}
For each motif, the motif-scaffolding method outputs a collection of 100 \emph{scaffolds}, each specified through the coordinates of backbone atoms (including those in the motif) of a single protein chain.
\begin{itemize}
\item{Coordinates for C$\alpha$ atoms must be given, and additional backbone atoms (N, C and O) may be specified as well.}
\item{All scaffolds must contain the same number of residues.  This fixed overall length of scaffolds is part of the problem specification.}
\item{Each scaffold is accompanied by metadata specifying the placement of the motif segment(s) in the scaffold sequence.
These placements may be chosen in any way (e.g.\ to match the order and spacing in the experimental structures, algorithmically, or manually).}
\end{itemize}

\paragraph{Evaluation:}
\begin{figure}[ht]
    \centering
    \includegraphics[width=1.0\textwidth]{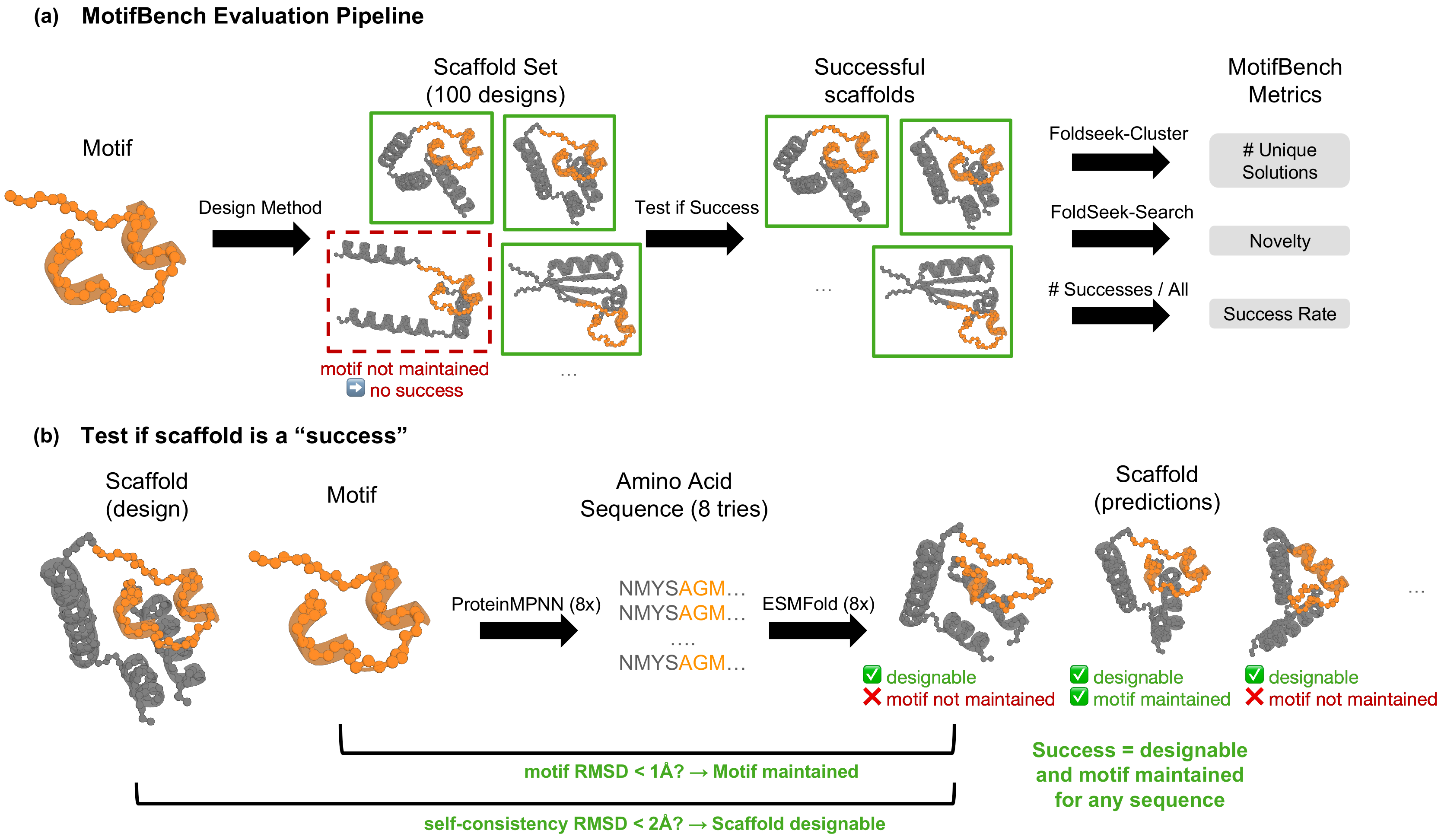} 
    \caption{MotifBench evaluation pipeline.}
    \label{fig:schematic} 
\end{figure}
Three metrics comprise the evaluation of each scaffold set: the \emph{number of unique solutions}, the \emph{novelty} of solutions, and overall \emph{success rate}.

The ``number of unique solutions'' counts the number of substantively distinct scaffolds in the set that are predicted to maintain the geometry of the motif within an experimentally realizable backbone.
This metric emphasizes that a method for the motif-scaffolding problem should ideally provide a variety of solutions;
protein design problems often have additional difficult-to-specify constraints and so benefit from further computational filtering and experimental screening.
This metric is computed as follows:

\begin{enumerate}
    \item{For each scaffold, eight amino acid sequences are generated using a fixed backbone sequence design method. MotifBench uses \texttt{ProteinMPNN} \citep{dauparas2022robust} as default. 
    If not all backbone atoms (N, C$\alpha$, C and O) are specified, then C$\alpha$-only \texttt{ProteinMPNN} is used.
    }
    \item{For each generated sequence, a backbone structure is predicted using \texttt{ESMFold} \citep{lin2022language}.}
    \item{A scaffold backbone is a ``success'' if at least one of the eight sequences satisfies both of the following creteria:
    \begin{itemize}
        \item {\textbf{Motif maintenance:} The root mean squared distance (RMSD) between the backbone atoms (N, C$\alpha$, and C) of the input motif and corresponding atoms of the predicted structure (the \emph{motifRMSD}) is at most 1.0 Angstrom (\AA);}
        \item{\textbf{Scaffold validity:} The RMSD between corresponding backbone atoms (C$\alpha$ only) of the generated and predicted structures (the \emph{self-consistency RMSD}, or \emph{scRMSD} \citep{trippe2022diffusion}) is at most 2.0 \AA.}
    \end{itemize}
    For each of the above steps, the Kabsch algorithm \citep{kabsch1976solution} is first used to align structures so that the minimum RMSD across all possible alignments is returned.
    }
    \item{The ``number of unique solutions'' is the number of clusters into which ``successful'' backbones are assigned by \texttt{Foldseek-Cluster} \citep{barrio2023clustering}.}
\end{enumerate}

Secondarily, a scaffold set is evaluated by two additional metrics:
\begin{itemize}
\item{``Novelty'' quantifies the typical distance of solutions to any structure in the PDB \citep{berman2002protein}. 
 To compute the novelty, for each success we use \texttt{FoldSeek-Search} \citep{van2024fast} to approximate the highest TM-score \citep{zhang2005tm} to any structure in the PDB.
 Then for each cluster of successes we compute the mean of one minus this TMscore within the cluster.
 The  ``Novelty'' is the mean of this score across clusters, or $0$ if there are no successful scaffolds.}
\item{The ``Success rate'' is the fraction of scaffolds in the set that are successes, independent of their diversity and novelty.}
\end{itemize}

\paragraph{The MotifBench score for ranking performance across a set of problems.}
To enable ranking the relative ability of a motif-scaffolding method to provide unique solutions across a collection of test cases we introduce the ``MotifBench score''.  The MotifBench score has the form

$$
\textrm{MotifBench score} = \frac{1}{\# \mathrm{\ test\ cases}} \sum_{i=1}^{\# \mathrm{\ test\ cases}} (100 + \alpha) \frac{\#\mathrm{\ unique \ solutions\ for\ case\ } i}{\alpha + \# \mathrm{\ unique \ solutions \ for \ case\ } i}\ , \quad \mathrm{with}\  \alpha=5.
$$

We choose the form of this score to capture that the marginal value of an additional solution for a given problem is much larger when the number of unique solutions is low.  If there were just one motif, one could rank on the number of unique solutions.
However, simply averaging this across misses that the marginal value of an additional solution is much larger when the number of solutions is low.

We note two properties of this metric:
\begin{itemize}
    \item It ranges from 0 to 100; the score is 0 if no solutions are found and 100 every backbone is a unique solution for every problem.
    \item The choice of $\alpha =5$ gives high weight to first solutions and increases much more slowly as the number of solutions grows larger.
\end{itemize}
As examples, with 1, 5, or 50 solutions the MotifBench score is 17.5, 52.5 or 95.5, respectively.

As a consequence of the second property, the MotifBench score reflects that it is preferable for a method to provide few solutions to more motifs rather than to provide more solutions to few motifs.
So a method (``method A'') that gives one unique solution for every test case would achieve a much higher score (17.5/100) than another method (``method B'') that returns one hundred solutions for one of the thirty total test cases and zero solutions to the twenty-nine every other problem (3.33/100).  By contrast, the simpler score that is just the average number of unique solutions would be lower for method A (1./100) than method B (3.33/100).

The MotifBench score is not, however, intended to capture every aspect of a motif-scaffolding method and cannot be taken to indicate that one method is superior than another in every setting.  Indeed a method that tends to provide a smaller number of unique solutions than another may still be preferable in some applications if the solutions it identifies are more novel or have different secondary structure content on average.

\paragraph{Pseudo-code algorithms detailing the evaluation procedure.}
\Cref{alg:test_backbone_set,alg:test_scaffold} detail the evaluation procedure and metrics. 
They use the following notation:
\begin{itemize}
\item{$\vec x_{n,l,a}^{\mathrm{design}}:$ 3D coordinate of atom $a$ in residue $l$ of designed scaffold $n$.}
\item{$I_{n, m}^{\mathrm{design}}:$ Index of the first residue of motif segment $m$ in scaffold $n$.  }
\item{$\vec x_{m,l, a}^{\mathrm{motif}}:$ 3D coordinate of atom $a$ in residue $l$ of motif segment $m.$} 
\item{$s_{m,l}^{\mathrm{motif}}:$ Amino acid type of residue $l$ of motif segment $m.$} 
\item{$r_{m,l}^{\mathrm{motif}}:$ Indicator of whether the amino acid type of residue $l$ of motif segment $m$ may be redesigned.}
\end{itemize}

\begin{algorithm}[t]
\caption{Compute metrics for a scaffold set for a given motif }\label{alg:test_backbone_set}
\begin{algorithmic}
\Procedure{backbone\_metrics}{$
\{\vec x_{n,l,a}^{\mathrm{design}} \},  
\{I_{n, m}^{\mathrm{design}}\}, 
\{\vec x_{m,l, a}^{\mathrm{motif}}\}, 
\{s_{m,l}^{\mathrm{motif}}\}, 
\{r_{m,l}^{\mathrm{motif}}\} 
$}
\LeftCommentOne{Identify successful scaffolds}
\State {$\textrm{success}_n
{\leftarrow} \texttt{testScaffold}(\{\vec x_{n,l,a}^{\mathrm{design} }\}, \{I_{n,m}^{\mathrm{design}}\}, \{x^{\mathrm{motif}}_{m,l, a}\}, \{s^{\mathrm{motif}}_{m,l}\}, \{r_{m,l}^{\mathrm{motif}}\})$}\Comment{$n\in \{1,\dots, 100\}$}
\State $\{\vec x_{n,l,a}^{\mathrm{success}}\} \gets \{ \{\vec x_{n, l, a}^{\mathrm{design}}\} \textrm{\ for\ } n\in \{1,\dots, 100\} \ \textrm{if success}_n=\mathbf{True} \}$  
\State $N_{\mathrm{success}} \gets | \{\vec x_{n,l,a}^{\mathrm{success}}\}|$
\State 
\LeftCommentOne{Cluster for the number of unique solutions.}
\State clusters $\gets \texttt{Foldseek-cluster}( \{\vec x_{n,l,a}^{\mathrm{success}}\}$)
\State num\_solutions = $|\textrm{clusters}|$
\State 
\LeftCommentOne{Compute novelty}
    \State $\{\mathrm{pdb\_tm}_{c,n}\} \gets \texttt{Foldseek-search}(\{x_{n,l,a}^{\mathrm{success}}\})$ 
    \Comment{$c\in \textrm{clusters}, n \in \textrm{c}$}
    \State {$\textrm{novelty}_c$} $\gets 1\ - \ \mathrm{mean}(\{\mathrm{pdb\_tm}_{c,n}\  \textrm{for}\  n \in c\}$) 
    \Comment{$c\in \textrm{clusters}$}
    \State {novelty} $\gets \mathrm{mean}(\{\mathrm{novelty}_c \ \textrm{for} \ c \in \textrm{clusters}\}$) 
\State 
\LeftCommentOne{Compute success rate}
\State $\textrm{success\_rate} = N_{\mathrm{success}} / 100$
\State \Return num\_solutions, novelty, success\_rate
\EndProcedure
\end{algorithmic}
\end{algorithm}

\begin{algorithm}
\caption{Test backbone for motif maintenance and scaffold validity}\label{alg:test_scaffold}
\begin{algorithmic}
\Procedure{testBackbone}{$\{\vec x_{l,a}^{\mathrm{design} }\},  \{I_{m}^{\mathrm{design}}\}, \{x^{\mathrm{motif}}_{m,l,a}\}, \{s^{\mathrm{motif}}_{m,l}\}, \{r_{m,l}^{\mathrm{motif}}\}$
}
\LeftCommentOne{Fixed-backbone sequence design with motif sequence restricted}
    \If{designed\_scaffold has N-Ca-C-O atoms}
        \Comment{Use full-backbone \texttt{ProteinMPNN}}
        \State{$\{s_{n,l}\} \leftarrow$ \texttt{ProteinMPNN}(
        $\{\vec x_{l,a}^{\mathrm{design} }\}, 
        \{I_{m}^{\mathrm{design}}\}, \{s^\mathrm{motif}_{m,l}\}, \{r_{m,l}^{\mathrm{motif}}\})$}
        \Comment {$n \in \{1, \dots, 8\}$}
        \Else
        \Comment{Use C$\alpha$ only \texttt{ProteinMPNN}}
        \State{$\{s_{n, l}\} \leftarrow \texttt{CA-ProteinMPNN}(
        \{\vec x_{l,C\alpha}^{\mathrm{design} }\}, 
        \{I_{m}^{\mathrm{design}}\},\{s^\mathrm{motif}_{m,l}\}, \{r_{m,l}^{\mathrm{motif}}\}$)}
        \Comment {$n \{1,\dots, 8\}$}
    \EndIf
    \LeftCommentOne{Structure prediction}
    \State $\{\vec x_{n,l,a}^{\mathrm{pred}}  \} \leftarrow $ \texttt{ESMFold}$(\{s_{n,l}\})$ \Comment {$n \in \{1,\dots, 8\}$}

    \State
    \LeftCommentOne{Motif maintenance: compute motif RMSD across segments (C, C$\alpha,$ and N atoms)}
    \State $\{\vec x_{n, m, l, a}^{\mathrm{motif}, \mathrm{pred}} \} \leftarrow 
    \{\vec x_{n,I_m+l, a}^{\mathrm{pred}}\}$
    \Comment{$n\in\{1,\dots, 8\}, m\in\{1, \dots\}, a\in \{N, C\alpha, C\}$}
    \State $\textrm{motifRMSD}_n$ $\leftarrow$ \texttt{ComputeAlignedRMSD}$(
    \{\vec x_{n, m, l, a}^{\mathrm{motif}, \mathrm{pred}}\},
    \{\vec x^{\mathrm{motif},\mathrm{true}}_{m,l, a}\}
    )$
        \Comment {$n \in \{1, \dots, 8\}$}
    \State
    
    \LeftCommentOne{Scaffold validity: compute self-consistency RMSD (C$\alpha$ only)}
    \State $\textrm{scRMSD}_n$ $\leftarrow$ \texttt{ComputeAlignedRMSD}$(
    \{\vec x_{n, l, C\alpha}^{ \mathrm{pred}}\},
    \{\vec x^{\mathrm{design}}_{n,l, C\alpha}\})$
        \Comment {$n \in \{1, \dots, 8\}$}
    \State
    \State successes $\leftarrow \{ n\  | \ \mathrm{motifRMSD}_n < 1$ \AA\  
    $ \mathbf{and}\ \mathrm{scRMSD}_n < 2$ \AA  \}
\If{successes =\{ \}}
    \Return  False
\EndIf
\State \Return  \textbf{True}
\EndProcedure
\end{algorithmic}
\end{algorithm}

\Cref{alg:test_scaffold} 
 relies on the Kabsch algorithm in two calls to \texttt{ComputedAlignedRMSD}.
 This algorithm is provided in \Cref{alg:compute_aligned_rmsd} of \Cref{sec:kabsch_alg}.

\section{The MotifBench test cases: composition and considerations
}\label{sec:benchmark_cases}

MotifBench comprises 30 test problems.
We specify these problems in \Cref{table:test_cases}, which has the following columns:
\begin{itemize}
\item{\textbf{PDB ID}: The Protein Data Bank identifier of the experimentally characterized structure from which the motif extracted.}
\item{\textbf{Group:} The problem group into which the motif is assigned. This grouping is defined based on the number of contiguous segments that comprise the motif: Group 1 motifs have only one segment, Group 2 motifs have 2 segments, and Group 3 motifs have 3 or more segments. We provide an additional visualization of motifs by group in \Cref{fig:group_1_problems,fig:group_2_problems,fig:group_3_problems}. }
\item{\textbf{Length}: The number of residues required in each scaffold.}
\item{\textbf{Motif Residues}: The chain ID and indices of residues that comprise the motif. Discontiguous residue ranges are separated by semicolons.}
\item{\textbf{Redesign Indices}: The indices of residues within motif segments for which the amino acid type is not constrained to match its identity in the reference protein, and will be ``redesigned'' during inverse-folding. This column is included because in cases where side-chain atoms are not involved in protein function, the motif-scaffolding problem may be made easier by allowing alternative amino acid types to be chosen for these positions during fixed-backbone sequence design.  This field is blank when no positions are allowed to be redesigned.}
\item{\textbf{Description:} A short explanation and reference for the source of the motif.}
\end{itemize}

\begin{footnotesize}
\begin{table}[htbp]
\footnotesize
\caption{MotifBench test-cases.}\label{table:test_cases} 
\begin{tabularx}{\textwidth}{|l|l|l|l|p{3.2cm}|p{3.3cm}|p{3.2cm}|}
\hline
\# & PDB  ID & Group & Length & Motif Residues & Positions where residue type may be designed & Description \\ \hline
\hline
1  & 1LDB & 1 & 125 & A186-206 & & Lactate dehydrogenase \citep{hayes2024simulating} \\ \hline
2  & 1ITU & 1 & 150 & A124-147 &  & Renal dipeptidase \citep{hayes2024simulating} \\ \hline
3  & 2CGA & 1 & 125  & A184-194	& & Strained chymotrypsinogen loop that undergoes conformation change \citep{wang1985bovine} \\ \hline
4  & 5WN9 & 1 & 75 & A170-189 & A170-175;A188-189 & RSV G-protein 2D10 site \citep{yang2021bottom} \\ \hline
5  & 5ZE9 & 1 & 100 & A229-243 & & P-loop \citep{hayes2024simulating} \\ \hline
6  & 6E6R & 1 & 75 & A25-35 & A25-35 & Ferredoxin Protein \citep{trippe2022diffusion} \\ \hline
7  & 6E6R & 1 & 200 & A25-35 & A25-35 & Ferredoxin Protein \citep{trippe2022diffusion} \\ \hline
8  & 7AD5 & 1 & 125 & A99-113 &  & Orphan protein \citep{wu2022omegafold} \\ \hline
9  & 7CG5 & 1 & 125 & A6-20 &  & Orphan protein \citep{wu2022omegafold} \\ \hline
10  & 7WRK & 1 & 125 & A80-94 & & Orphan protein \citep{wu2022omegafold} \\ \hline
\hline
11  & 3TQB & 2 & 125 & A37-51;A65-79 & & Parallel beta strand and loop \citep{tkaczuk2013structural} \\ \hline
12  & 4JHW & 2 & 100 & F63-69;F196-212 & F63;F69;F196;F198; F203;F211-212 & RSV F-protein Site 0 \citep{sesterhenn2020novo} \\ \hline
13  & 4JHW & 2 & 200 & F63-69;F196-212 & F63;F69;F196;F198; F203;F211-212  & RSV F-protein Site 0 \citep{sesterhenn2020novo} \\ \hline
14  & 5IUS & 2 & 100 & A63-82;A119-140 & A63;A65;A67;A69;\ A71-72;A76;A79-80;A82;A119-123;\ A125;A127;A129-130;\ A133;A135;A137-138;A140 & PD-L1 binding interface on PD-1 \citep{watson2022broadly} \\ \hline
15  & 7A8S & 2 & 100 & A41-55;A72-86 &  & Orphan protein \citep{wu2022omegafold} \\ \hline
16  & 7BNY & 2 & 125 & A83-97;A111-125 &  & Orphan protein \citep{wu2022omegafold} \\ \hline
17  & 7DGW & 2 & 125 & A22-36;A70-84 &  & Orphan protein \citep{wu2022omegafold} \\ \hline
18  & 7MQQ & 2 & 100 & A80-94;A115-129 & & Orphan protein \citep{wu2022omegafold} \\ \hline
19  & 7MQQ & 2 & 200 & A80-94;A115-129 &  & Orphan protein \citep{wu2022omegafold} \\ \hline
20  & 7UWL & 2 &  175 & E63-73;E101-111 & E63-73;E101-103;E105-111 & IL17-RA interface to IL17-RB \citep{wilson2022organizing} \\ \hline
\hline
21  & 1B73 & 3 & 125 & A7-8;A70;A178-180 & A179 & Glutamate racemase active site \citep{ribeiro2018mechanism} \\ \hline
22  & 1BCF & 3 & 125 & A18-25;A47-54;A92-99;A123-130  & A19-25;A47-50;A52-53; A92-93;A95-99;A123-126;A128-129 & Di-iron binding motif \citep{watson2022broadly} \\ \hline
23  & 1MPY & 3 & 125 & A153;A199;A214;A246;\ A255;A265 & & Catechol deoxygenase active site \citep{ribeiro2018mechanism} \\ \hline
24  & 1QY3 & 3 & 225 & A58-71;A96;A222  & A58-61;A63-64;A68-71  & GFP pre-cyclized state \citep{hayes2024simulating}.\footnote{The deposited structure includes an inactivating mutation (R96A) mutation.  Following \citep{hayes2024simulating}, this mutation must be reverted in generated scaffolds.} \\ \hline
25  & 2RKX & 3 & 225 & A9-11;A48-50;A101;A128;\ A169;A176;A201;A222-224 & A10;A49;A223 & De novo designed Kemp eliminase \citep{rothlisberger2008kemp} \\ \hline
26  & 3B5V & 3 & 200 & A51-53;A81;A110;A131;\ A159;A180-184;A210-211;A231-233 & A52;A181;A183;A232 & De novo designed retro-aldol enzyme \citep{jiang2008novo} \\ \hline
27  & 4XOJ & 3 & 150 & A55;A99;A190-192 & A191  & Trypsin catalytic triad and oxyanion hole \citep{du2024conformational} \\ \hline
28  & 5YUI & 3 & 75 & A93-97;A118-120;A198-200 & A93;A95;A97;A118;A120  & Carbonic anhydrase active site \citep{watson2022broadly} \\ \hline
29  & 6CPA & 3 & 200 & A69-72;A127;A196;\ A248;A270 & A70-71 & Carboxypeptidase active site \citep{ribeiro2018mechanism} \\ \hline
30  & 7UWL & 3 & 175 & E63-73;E101-111;E132-142;E165-174 & E63-73;E101-103;E105-111;E132-142;E165-174 & IL17-RA interface to IL17-RB \citep{wilson2022organizing} \\ \hline
\end{tabularx}
\end{table}
\end{footnotesize}

These test cases are derived from several sources.
\begin{itemize}
    \item{Eight problems are motifs considered in published protein design papers collected in an earlier benchmark set \citep[Table S9 of ][]{watson2022broadly}. Compared to the 25 problems in the this earlier benchmark, we drop one problem that involves multiple chains (6VW1), two problems for which we observed ESMFold to dramatically increase the success rate compared to AlphaFold2 with no MSA input (1QJG, 1PRW), and several problems that are already readily solved by existing methods.  The problems we include this set are 5WN9 (4), 6E6R (6 and 7), 4JHW (12 and 13), 5IUS (14), 1BCF (22), and 5YUI (28).}
\item Eight problems are fragments of ``orphan'' proteins selected in \citep[Table S11 of ][]{watson2022broadly} from structures in \citep{wu2022omegafold} with little sequence or structure homology other known proteins.  These were selected from among an initial set of 25 problems for structural diversity and greater difficulty.  These problems are 7AD5 (8), 7CG5 (9), 7WRK (10), 78AS (15), 7BNY (16), 7DGW(17), 7MQQ (18-19).
\item{Four problems are obtained from the test cases considered in another protein design paper \citep{hayes2024simulating}.  These problems are 1LDB (1), 1ITU (2), 5ZE9 (5), and 1QY3 (25).  Problem the PDB entry 1QY3 is crystal structure of the green fluorescent protein (GFP) with a mutation (Arg $\rightarrow$ Ala at residue 96) which prevents the formation of the GFP fluorophore.  Our motif definition involves the reversion of this mutation to the native Arginine.}
\item{One problem, 2CGA (3) was identified as a difficult single-segment case in a strained conformation; this motif is loop in chymotrypsinogen that has a documented conformation change after a cleavage of its native scaffold and activation to chymotrypsin \citep{wang1985bovine}.}
\item{One problem, 3TQB (11) was selected because it includes a parallel beta strand structure not elsewhere represented in the benchmark \citep{tkaczuk2013structural}.}
\item{Six problems are additional enzyme active sites: 
\begin{itemize}
    \item{Three are from natural enzymes in the ``Mechanism and Catalytic Site Atlas'' \citep{ribeiro2018mechanism}: 1B73 (21), 1MPY (23) and 6CPA (29). }
    \item{Two are active sites of \emph{de novo} designed enzymes, 2RKX (22) \citep{rothlisberger2008kemp} and 3BV5 (26) \citep{jiang2008novo}.  }
    \item{One is the active residues of a serine protease described in a structural study \citep{du2024conformational}: 4XOJ (27).}
\end{itemize}
For the motifs from natural enzymes, the motif residues are chosen from among those that are documented as involved in the catalytic mechanism and or can be observed from the experimental structure to make polar contacts within the annotated active site.  When there is a gap between involved residues of no more than three amino acids in the sequence these residues are also included as part of the motif, but with these positions marked as redesignable.  
Residues are also marked as redesignable when only their backbone atoms appear to be involved in the mechanism.
For the motifs from \emph{de novo} design papers, the motif residues are chosen to be those that were chosen when constructing the putative active site.
 }
\item {Two problems, 7UWL (20 and 30), are segments of a binding interface of IL17-RA that interacts with IL17-RB \citep{wilson2022organizing}.
The native interface involves four segments; two of these segments are included in problem 20 and all four segments are included in problem 30.
These problems were chosen for their difficulty and the potential therapeutic relevance of novel scaffolds that reconstitute this interface.}

\end{itemize}

We prioritized the following characteristics when selecting the motifs above.
\begin{itemize}
    \item{\textbf{Relevance to design}: Most of the test cases are minimal, biochemically active substructures obtained from or characteristic of protein design problems.}
    \item{\textbf{Diversity}: The motifs in the benchmark exhibit a range of characteristics:
    \begin{itemize}
        \item The number of residues in the motifs ranges from as few as five to several dozens.  Because the lengths demanded in a scaffold can be a significant factor for the performance of some methods, for three motifs (4JHW, 6E6R, and 7MQQ) we include two problem variations with different scaffold lengths.
        \item The number of contiguous segments ranges from one to eight.
        \item The number of residues demanded of scaffolds varies from 75 to 225. 
        \item The secondary structure of motifs across problems includes helical, strand, and loop segments, and combinations thereof.
        \end{itemize}}
\item{\textbf{Inclusion of ``Orphan'' motifs}:  Eight ``orphan protein'' motifs are included to ease assessment of possible dependence of measured performance on overfitting by data-driven motif-scaffolding methods and the MotifBench evaluation pipeline.  In particular, evaluations may be artificially inflated for motifs that are structurally conserved (and therefore more highly represented in protein sequence databases and in the PDB) if the corresponding native sequence and then native motif structure are readily predicted by ProteinMPNN and ESMFold as a result of memorization even without a suitable supporting scaffold.  By contrast, motifs extracted from orphan proteins are less likely to be highly represented in these datasets and therefore are less likely to be susceptible to this bias.} 
\end{itemize}

\section{Baseline performance and analysis of reference scaffolds}
\label{section:baseline_section}
We demonstrate MotifBench using the well-established motif-scaffolding method RFdiffusion \citep{watson2022broadly} to assess various aspects of the evaluation pipeline. Key ingredients are summarized below:
\begin{itemize}
    \item \textbf{Performance}: We provide an running example for practitioners to start with MotifBench and demonstrated the challenge of designing cases in MotifBench.
    \item \textbf{Stability}: We showcased MotifBench is robust to the randomness from multiple replicates of both design and evaluation is on a low level.
    \item \textbf{Sensitivity to forward folding method}: We demonstrate a relatively low sensitivity to the choice of two commonly-used forward folding methods, ESMFold and AlphaFold2 (no MSA input), for validating designed scaffolds.
    \item \textbf{Rationality of curated cases}: We show that there exist reasonable solutions for curated cases in MotifBench by running evaluation on reference proteins from which the motifs were defined.
\end{itemize}

Altogether, these findings (1) suggest that substantial improvements could be made to motif-scaffolding methods and (2) point to limitations of the MotifBench V0 evaluation that might be addressed by future versions.

\paragraph{Demonstration of MotifBench and performance evaluation with RFdiffusion scaffolds.}

To assess the feasibility and difficulty of the MotifBench test cases, we evaluated scaffold sets produced with RFdiffusion \citep{watson2022broadly}. We chose RFdiffusion for both its popularity among motif-scaffolding methods and its good performance relative to more recent motif-scaffolding methods \citep[see e.g.][]{wu2024practical,lin2024out,yim2024improved}.

We generated 100 scaffolds for each of the 30 test cases using the open-source RFDiffusion implementation with default hyperparameter settings and contigs under the specification of MotifBench.  Generation required approximately 30 GPU hours across a variety of GPU types on a university cluster. We then evaluated these scaffolds with MotifBench. We provide full details of this generation and code to replicate generation of the scaffolds sets at \url{github.com/blt2114/motif_scaffolding/benchmark/example} and have uploaded the scaffolds and associated metadata in the format required by MotifBench accompanied with all evaluated results to zenodo at \url{https://zenodo.org/records/14731790} for replicability.

RFdiffusion provided at least one solution on 16 of the 30 cases, which indicates greater difficulty of MotifBench test cases as compared to earlier the benchmark set introduced in \citep{watson2022broadly};
on this earlier benchmark set, RFdiffusion has been found by \citet{watson2022broadly, yim2024improved,zhang2023protein} to provide at least one solution in 20 of 24 single-chain test cases.  Across the present test cases, the mean number of unique solutions across problems was 8.83, but 5 or more unique solutions were found for only 7 test cases. The mean novelty across cases was 0.19 and the overall MotifBench score was 28.05. RFdiffusion identified solutions for 7/10 cases for cases in both group 1 and group 2 but 2/10 for group 3 motifs, highlighting challenges for scaffolding motifs including multiple discontinuous segments (\Cref{table:RFdiff_replicates}).

\paragraph{Stability against stochasticity in MotifBench evaluation and scaffold generation.} A challenge in motif-scaffolding evaluation is stochasticity arising from (1) the sequence design step of the evaluation pipeline and (2) the scaffold generation. To assess the stability and reproducibility of MotifBench metrics, we evaluate the variability of MotifBench results across replicates due to each source of stochasticity:
{\begin{itemize}
    \item \textbf{Stochasticity across evaluations on a single scaffold set}. We repeated the evaluation on the initial set of the generated scaffold described above four times. See ``Variability from Benchmark'' in \Cref{table:RFdiff_replicates}.
    \item \textbf{stochasticity across scaffold generations}. We also replicated the RFdiffusion scaffold generation procedure and evaluated thenceforth. See ``Variability from Scaffolds'' in \Cref{table:RFdiff_replicates}.
\end{itemize}}

We observed variability from both sources, with greater variability across replicate scaffold generations in which case both sources of stochasticity contribute. The standard deviation of MotifBench score was 0.39 for multiple evaluations and 0.47 for multiple scaffold sets. Rougnly, this result suggests that differences in the MotifBench score of less than about 1 may not be sufficient to confidently conclude one method outperforms another based on a single evaluation.

\paragraph{Sensitivity of MotifBench to different structure prediction methods.} We explored how the choice of structure prediction method (ESMFold vs. AlphaFold2 without MSA input) affects MotifBench results. While performance was similar for most cases, we observed notable variability for a few cases between the two methods (\Cref{table:RFdiff_folding_method}); this discrepancy has a non-trivial impact on the MotifBench score (28.1 with ESMFold vs. 22.5 with AF2), yet does not imply superiority of one method over the other, as it may reflect either false positives from ESMFold or false negatives from AlphaFold2 (in non-MSA mode) or both.

\paragraph{Evaluation of reference scaffolds from experimental structures and the feasibility of ``unsolved'' problems.}
We next consider the extent to which the failure of RFdiffusion to identify solutions for 14/30 test cases owes to limitations in scaffold generation that could be addressed by improved methods versus limitations of the MotifBench evaluation; though each motif comes from an experimentally characterized structure, it remains possible that neither this experimental \emph{reference} scaffold nor any other scaffold could pass the MotifBench evaluation.  To assess these possibilities we evaluated the reference scaffold for each motif according to MotifBench (\Cref{table:reference_RFdiff}) and compare the overlaps in the test cases for which RFdiffusion succeeds or fails with the cases for which the reference scaffold succeeds or fails in a contingency table (\Cref{table:reference_RFdiff_contingency}).

The reference scaffold was found to be a ``success'' in the majority (20/30) cases (\Cref{table:reference_RFdiff}).  RFdiffusion failed to identify a solution for 6 of these successes, indicating the low success rates are at least partly due to design limitations rather than evaluation. Surprisingly, 2 of these 6 cases for which the reference scaffold is a success but RFdiffusion identifies no solutions are motifs derived from \textit{de novo} proteins that were scaffolded by a pre-deep learning method, 2RKX \citep{rothlisberger2008kemp} and 3B5B \citep{jiang2008novo}.  This result indicates the recent wave of deep-learning methods may not improve uniformly upon techniques used for case-specific designs more than 15 years ago.  

For the remaining 8 unsolved cases the reference scaffold fails to pass the MotifBench evaluation.
Upon inspection, the majority of these failure cases exhibit a high proportion of loop regions, suggesting a limitation of the evaluation pipeline to precisely reconstruct flexible regions in these motifs.
This limitation could owe to the sequence design or structure prediction steps, and may be addressed by improvements in methodologies for these tasks, and could be incorporated into future versions of MotifBench.
However, this failure is not conclusive evidence for the impossibility of identifying successes for these cases even with the MotifBench evaluation; indeed, for 2 of the cases for which the reference scaffold is not a success RFdiffusion nonetheless identifies passing scaffolds.

\section{Benchmark implementation details}
Evaluation scripts implementing the benchmark are provided at \url{https://github.com/blt2114/MotifBench} along with detailed instructions for formatting the inputs, i.e. the scaffold pdb-files and metadata.
We here briefly discuss compute requirements and other software upon which our benchmark implementation is built.

\paragraph{Compute requirements:}
Due to the heavy use of neural network methods,
evaluation of the full benchmark benefits significantly from GPU acceleration.
The benchmark takes roughly 36 hours on one Nvidia A4000 GPU.
For fast debugging purposes, we recommend running for a single problem;
for example on problem 6 (6E6R with 75 residue scaffold) evaluation requires only a few minutes.

\paragraph{Code sources:}
Beyond the tools used for evaluation already mentioned, this projects adapts code from several other open-source projects.  These include Scaffold-Lab \citep{zheng2024scaffold}, FrameDiff \citep{yim2023se},
the RigidTransform3D \citep{nghiaho12_rigid_transform_3D} implementation of the Kabsch algorithm \citep{kabsch1976solution,arun1987least}, and Openfold \citep{ahdritz2024openfold}.
\section{Community guidelines}
This section describes best practices for method development that we suggest to
those using MotifBench to improve reproducibility and community progress.
We encourage discussion about this benchmark in the form of issues on the github repository
so that the benchmark may be usefully updated in the future.

\paragraph{Reproducibility:}
Method developers publishing results using the benchmark are encouraged to make their code available so that others may replicate their results.
In the absence of code, the designed backbones can be shared.
Designed scaffold structures for the entire benchmark should demand roughly 100Mb.
Such files may be shared publicly through open data platforms such as Zenodo \citep{zenodo} or the Open Science Framework (OSF) \citep{foster2017open}.

\paragraph{Compute time:}
When reporting results computed using this benchmark,
please also report the compute expense of generating the backbones on which the benchmark is evaluated.

\paragraph{Problem-specific adjustments:}
It is ``okay'' to tailor methods to each specific problem ({e.g.} choices of placement of motif segments or method-specific hyperparameters).
However, problem specific adjustments should be noted and explained to the extent that they are necessary for reproducibility.

\section*{Acknowledgements}
We thank Daniel J. Diaz for suggestions on enzyme active site test cases (1B73, 1MPY, and 6CPA) and Steven C. Wilson for the IL17-RA binding interface motif (7UWL).
We thank Bozitao Zhong, Ting Wei, Kexin Liu, Junjie Zhu for helpful discussions. Z.Z., B.Z., and H.C. acknowledge the supported by the Center for HPC at Shanghai Jiao Tong University, and the National Key Research and Development Program of China (2020YFA0907700 and 2023YFF1205102), the National Natural Science Foundation of China (21977068 and 32171242), and the Fundamental Research Funds for the Central Universities (YG2023LC03).

\pagebreak

\bibliography{references}

\newpage
\appendix
\addtocontents{toc}{\protect\setcounter{tocdepth}{-1}}
\appendixpage

This appendix is organized as follows.
\Cref{sec:considerations} provides a discussion of choices incorporated into the MotifBench evaluation approach beyond those described in the main text.
\Cref{sec:kabsch_alg} provides the algorithm for computing aligned error that is part of the evaluation pipeline but which was deferred from \Cref{sec:task_specification}.
\Cref{sec:motif_vis} provides visualizations of each of the test cases and an problem specification file.
Finally, \Cref{sec:demo} provides full results of the evaluations described in \Cref{section:baseline_section}.

\section{Considerations and subjective choices in MotifBench evaluation}\label{sec:considerations}
Several steps of the pipeline involve necessarily subjective choices made in order to precisely define replicable evaluation procedures and metrics.
Pending community feedback, these choices may be altered in subsequent MotifBench versions.

\subsection*{Problem specification, inputs, and outputs}
\paragraph{Placement of motif segments:}
Most previous evaluations have relied on pre-specified orderings and suggested ranges for the placements of sequence-contiguous motif segments within scaffolds (see, e.g.\ the  ``contig'' specifications in \citep{watson2022broadly}).
Such pre-specification simplifies the task and allows methods for image inpainting/outpainting to be more easily applied, as these methods typically assume a fixed mask.
Accordingly, we include as suggested motif-placement for each problem chosen as the placement of the motif within the experimentally validated scaffold in the PDB from which the motif is derived.

However, ideal placements of motif segments may not be known in general for realistic problems and so it may be beneficial to automate this choice.
Furthermore, it may be helpful to vary this hyper-parameter to achieve larger numbers of diverse solutions,
success rates, and novelty (e.g.\ by random sampling \citep[see e.g.][]{watson2022broadly} or dynamically throughout the course of generation \citep[see e.g.][]{wu2024practical}).

\paragraph{Scaffold length:}
By contrast with the placement of motif segments, the lengths of the scaffolds to be generated are specified as part of each problem.
The scaffold length is also a subjective hyper-parameter whose choice could be automated in principle.
However, we restrict to a single length to avoid length-dependent biases that can appear in several parts of the evaluation pipeline, e.g. diversity and novelty.
In the problem specifications in \cref{table:test_cases}, these scaffold lengths are subjectively chosen based on the length of the native scaffolds and what seemed plausible to the authors.

\paragraph{Number of sequences per backbone:}
Most previous evaluations have relied on the design of several sequences for each backbone and evaluated generations as successes if the metrics of one or more of the generated sequences passed the cutoffs.
This choice reflects that computational filtering is a practical technique that is commonly in protein engineering,
and addresses the limitation that the stochastic generations of common fixed backbone sequence design methods sometimes fail to identify an adequate sequence on the first generation.
However using a larger number of sequences also increases computational cost and risks higher false-positive rates.
MotifBench specifies eight sequences for each scaffold as a practical balance that has been used across several previous works \citep{trippe2022diffusion,yim2023se,watson2022broadly}.

\subsection*{Choices of thresholds in evaluation}
We next discuss several subjective thresholds in the definition of evaluation metrics.

\paragraph{motifRMSD and scRMSD thresholds:}
The success criteria in MotifBench includes thresholds on the precision to which the motif and full scaffold must be recapitulated.
The 1{\AA} threshold on motif recapitulation is chosen to demand atomic precision;
by comparison, the atomic radius of a hydrogen atom is about 1.2{\AA}.
The 2{\AA} threshold on backbone recapitulation is set as a coarser level of precision that demands the overall backbone structure can be designed. 

\paragraph{Structural similarity thresholds for clustering and novelty:}
We adopt default settings of Foldseek-search and Foldseek-cluster for simplicity.
However, these methods have several parameters that impact their behavior for which different settings could in principle better align the number of unique solutions and novelty metrics with the efficacy of a design method in application.

\paragraph{Filtering on structure prediction confidence:}
Predictions of high confidence in the accuracy of protein structure prediction outputs has been included in previous motif-scaffolding evaluations as a proxy for the quality of designed scaffolds.
However, this criterion is partly largely redundant with the designability metric of scRMSD$<2${\AA} and is highly dependent on the accuracy of confidence head for specified structure prediction method.
Therefore, we dropped this criterion for simplicity.

\subsection*{Software choices}
MotifBench makes several choices of open sources methods in the evaluation pipeline on the basis of 
demonstrated predictive power for viability in \emph{in vitro} experiments,
computational efficiency and convenience,
and prior community adoption.

\paragraph{Fixed-backbone sequence design (inverse folding) method:}
A number of inverse folding methods exist that could in principle be used for the the sequence design step \citep[e.g.][]{dauparas2022robust,liu2022rotamer,anand2022protein}.
MotifBench specifies ProteinMPNN \citep{dauparas2022robust} with default parameters (including sampling temperature set to 0.1) for its precedent in past work \citep[e.g.][]{trippe2022diffusion}, its significant experimental validation \citep{dauparas2022robust,watson2022broadly}, and for its familiarity to the authors.

\paragraph{Structure prediction method:}
Several public software packages provide an accurate prediction of a protein structure from its amino acid sequence \citep{baek2021accurate,jumper2021highly,lin2022language,wu2022omegafold}. 
MotifBench specifies folding using ESMFold \citep{lin2022language} for its simplicity of implementation and computational efficiency.
However, the predictions of any computational structure prediction method may be incorrect and should be interpreted with this fact in mind;
and in some cases \texttt{ESMFold} is known to be less predictive of experimental viability as compared to AlphaFold2 \citep{jumper2021highly} when run with no multiple sequence alignment input \citep{martin2023validation}.
As such, the choice of ESMFold in MotifBench should not be interpreted as an endorsement for its use as an \emph{in silico} filter in protein design campaigns and we encourage users to report results by both ESMFold and AlphaFold2 (with no MSA input) for an orthogonal test when not computationally burdensome.

\paragraph{Structural clustering and similarity method:}
MotifBench specifies the use of \texttt{Foldseek-cluster} \citep{barrio2023clustering} and \texttt{Foldseek-search} \citep{van2024fast} for clustering and novelty evaluation (version \texttt{8.ef4e960}).
We choose these approaches as a alternative to the more widely used TMscore \citep{zhang2005tm} for their computational speed.

A limitation of Foldseek-Cluster is that it sometimes raises opaque errors during a ``prefilter'' step on certain scaffold sets.
We found that this error can be resolved by (1) adding an additional backbone unrelated to the design task to scaffold set, (2) re-running Foldseek-Cluster on the augmented set, and (3) removing the unrelated protein from the clustering results to limit its impact on the \emph{number of unique solutions} metric.

\section{Algorithmic details of aligned error computation}\label{sec:kabsch_alg}

\begin{algorithm}
\caption{Compute aligned root mean squared distance via Kabsch
Algorithm}\label{alg:compute_aligned_rmsd}
\begin{algorithmic}
\Procedure{ComputeAlignedRMSD}{$\{\vec x_{l}^{\mathrm{pred}}\}, \{\vec x_{l}^{\mathrm{design}}\}$}
    \LeftCommentOne{Center lists of coordinates by subtracting centers of mass}
    \State $\{\vec{x}_l^{\mathrm{pred}}\} \leftarrow \{\vec{x}^{\mathrm{pred}}_{l} -\frac{1}{L}\sum_{l^\prime=1}^{L} \vec{x}^{\mathrm{pred}}_{l^\prime}\}$
        \Comment {$l \in \{1, \dots, L\}$}
    \State $\{\vec{x}_l^{\mathrm{design}}\} \leftarrow \{\vec{x}^{\mathrm{design}}_{l} - \frac{1}{L}\sum_{l^\prime=1}^{L} \vec{x}^{\mathrm{design}}_{l^\prime}\}$
        \Comment {$l \in \{1, \dots, L\}$}
    
    
    \State
    \LeftCommentOne{Compute singular value decomposition of cross-covariance matrix}
    \State $U\Sigma V^T \leftarrow \texttt{SVD}( \sum_{l} \vec{x}_l^{\mathrm{pred}}(\vec{x}_l^{\mathrm{design}})^T)$
    
    \State
    \LeftCommentOne{If $UV^T$ is not a rotation, apply a reflection along direction with least singular value}
    \State $d \leftarrow \text{sign}(\det(UV^T))$
    \State $S \leftarrow \begin{pmatrix} 1 & 0 & 0 \\ 0 & 1 & 0 \\ 0 & 0 & d \end{pmatrix}$
    
    \State
    \LeftCommentOne{Compute and apply optimal rotation matrix}
    \State $R \leftarrow VS U^T$
    \State $\{\vec{x}_l^{\mathrm{aligned}}\} \leftarrow \{R\vec{x}_l^{\mathrm{pred}}\}$
        \Comment {$l \in \{1, \dots, L\}$}
    
    \State
    \LeftCommentOne{Compute the square root of the mean squared distance (RMSD}
    \State $\text{RMSD} \leftarrow \sqrt{\frac{1}{L}\sum_{l=1}^{L} \|\vec{x}^{\mathrm{aligned}}_{l} - \vec{x}^{\mathrm{design}}_{l}\|^2}$
    
    \Return RMSD
\EndProcedure
\end{algorithmic}
\end{algorithm}

\Cref{alg:test_scaffold} relies on the 
Kabsch algorithm to compute the minimum aligned root mean squared distance between atomic structures.
We detail this computation in \Cref{alg:compute_aligned_rmsd}.

\section{Test case visualizations and example specification}\label{sec:motif_vis}
\paragraph{Visualization:} We divide the benchmark set into three groups: Group 1 for single-segment motifs, Group 2 for double-segment motifs and Group 3 for multiple-segment motifs, with 10 problems for each group. We visualize the motif problems in \Cref{fig:group_1_problems,fig:group_2_problems,fig:group_3_problems}.

\paragraph{Example motif specification PDB:}  We provide an example motif specification in \Cref{fig:example_motif}.

\clearpage 
\pagebreak

{
\begin{figure}[ht]
    \centering
    \includegraphics[width=1.0\textwidth]{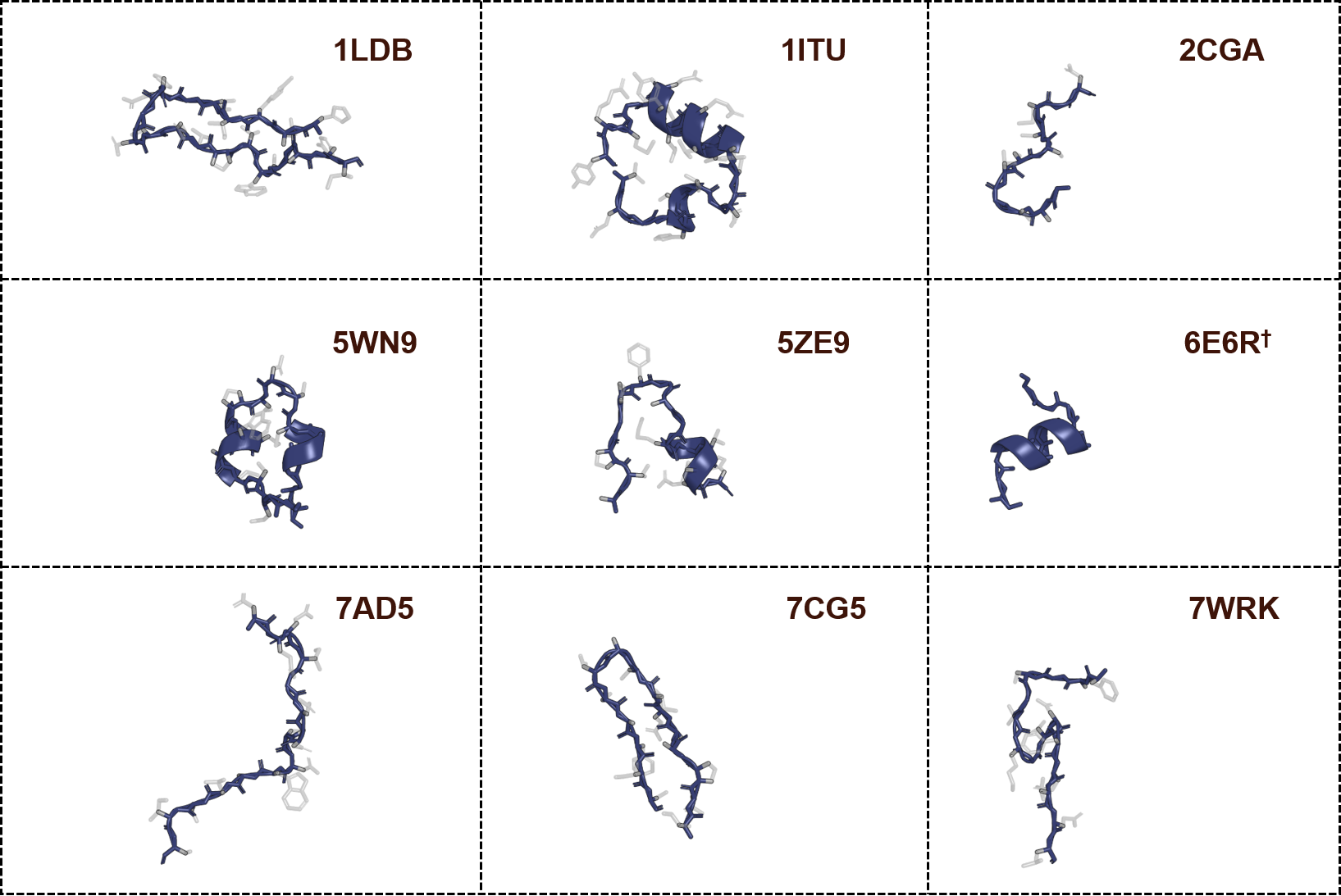} 
    \caption{Visualization of single-segment motifs in PyMol.  Colored by segment.  Side-chain atoms are shown only for positions for which amino acid type is fixed in the problem specification. ``\dag''  denotes the problem has a variation case with a different specified scaffold length.}
    \label{fig:group_1_problems} 
\end{figure}
}

\clearpage 
\pagebreak

{
\begin{figure}[ht]
    \centering
    \includegraphics[width=1.0\textwidth]{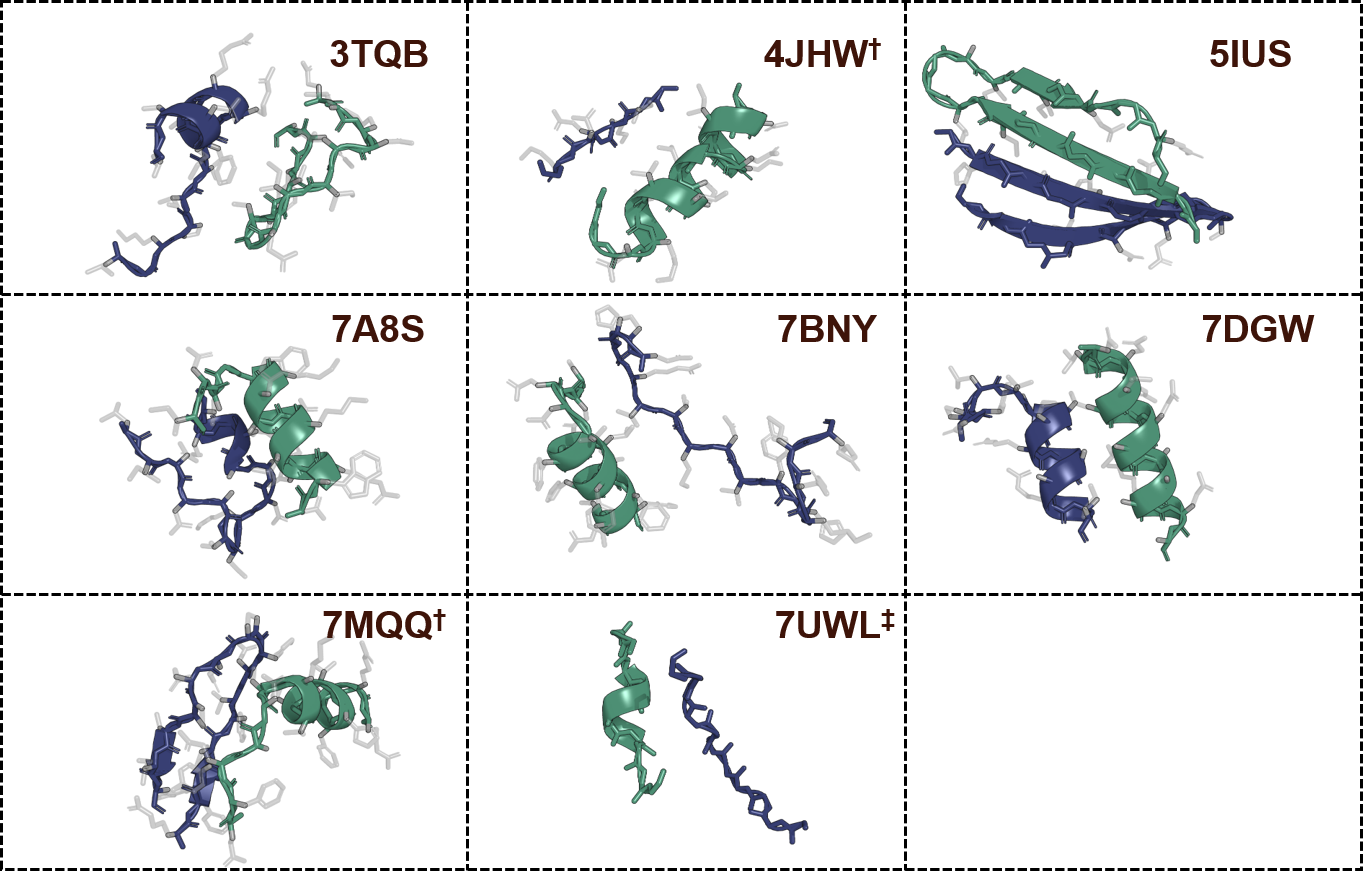} 
    \caption{Visualization of double-segment motifs in PyMol. Colored by segment.  Side-chain atoms are shown only for positions for which amino acid type is fixed in the problem specification. ``\dag''  denotes the problem has a variation case with a different specified scaffold length. ``\ddag'' denotes the problem has a variation case with different specified motif segments from the same reference protein.}
    \label{fig:group_2_problems} 
\end{figure}
}

\clearpage 
\pagebreak

{
\begin{figure}[ht]
    \centering
    \includegraphics[width=1.0\textwidth]{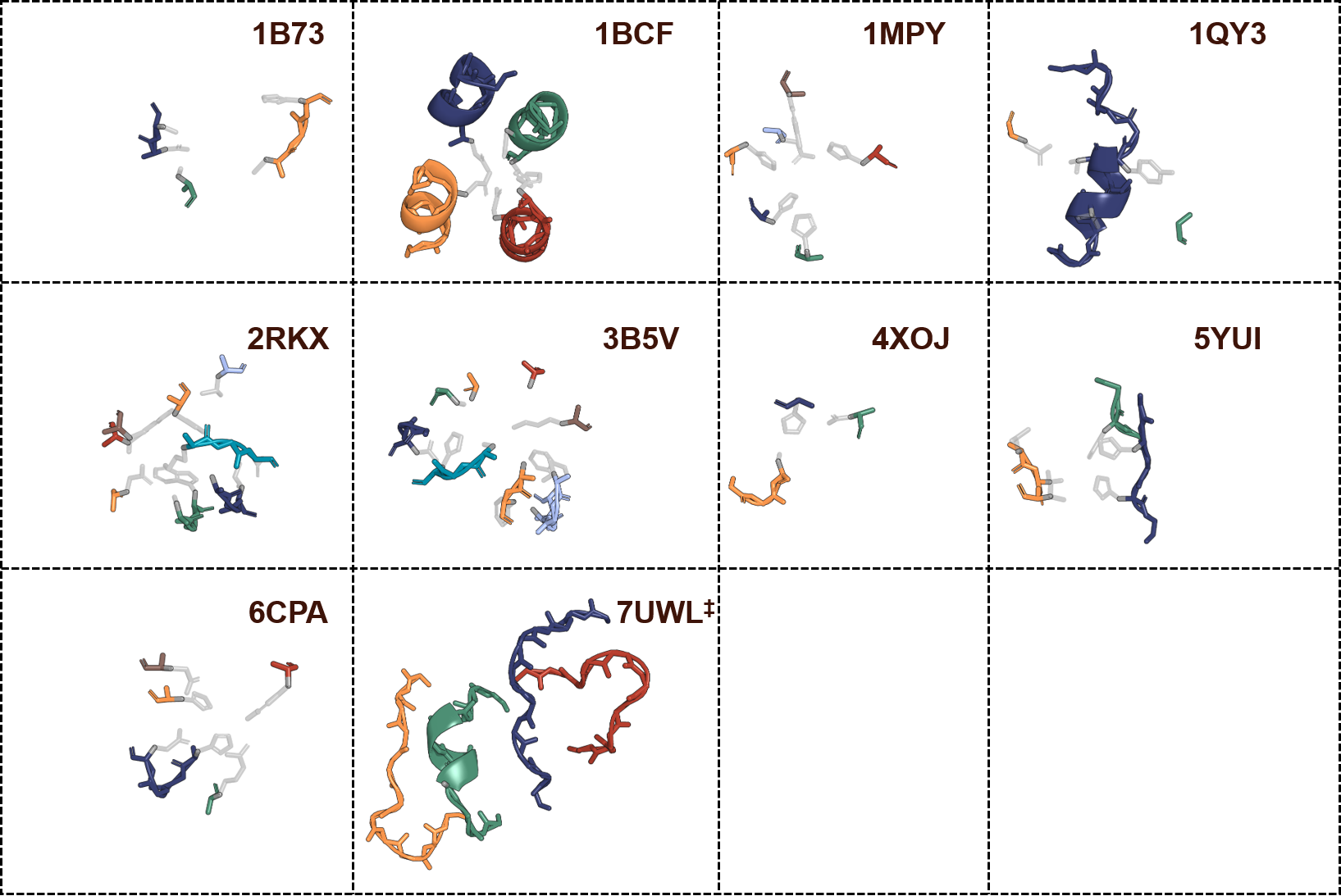} 
    \caption{Visualization of multiple-segment motifs in PyMol. Colored by segment.  Side-chain atoms are shown only for positions for which amino acid type is fixed in the problem specification. ``\ddag'' denotes the problem has a variation case with different specified motif segments from the same reference protein.}
    \label{fig:group_3_problems} 
\end{figure}
}

\clearpage 
\pagebreak

{

\begin{figure}
    \centering
\begin{tcolorbox}[
  title={},
  fonttitle=\bfseries,       
  colback=white,             
  colframe=black,            
  coltitle=white,            
  sharp corners,             
  boxrule=0.5mm,             
  toptitle=0.2mm,              
  bottomtitle=1mm,           
  titlerule=1mm              
]

\small 
\begin{verbatim}
REMARK 1 Reference PDB ID: 4XOJ
REMARK 2 Motif Segment Placement in Reference PDB: 39;A;43;B;90;C;46
REMARK 3 Length for Designed Scaffolds: 150
ATOM      1  N   HIS A   1      -2.924  -3.724   2.088  1.00  0.00           N  
ATOM      2  CA  HIS A   1      -1.871  -3.781   3.096  1.00  0.00           C  
ATOM      3  C   HIS A   1      -1.802  -2.517   3.945  1.00  0.00           C  
ATOM      4  O   HIS A   1      -1.071  -2.501   4.936  1.00  0.00           O  
ATOM      5  CB  HIS A   1      -0.502  -4.109   2.485  1.00  0.00           C  
ATOM      6  CG  HIS A   1       0.119  -2.995   1.722  1.00  0.00           C  
ATOM      7  ND1 HIS A   1       0.033  -2.839   0.365  1.00  0.00           N  
ATOM      8  CD2 HIS A   1       0.846  -1.950   2.150  1.00  0.00           C  
ATOM      9  CE1 HIS A   1       0.703  -1.723   0.052  1.00  0.00           C  
ATOM     10  NE2 HIS A   1       1.215  -1.148   1.102  1.00  0.00           N  
TER
ATOM     11  N   ASP B   1      -4.487  -6.677  -3.743  1.00  0.00           N  
ATOM     12  CA  ASP B   1      -4.063  -5.623  -2.793  1.00  0.00           C  
ATOM     13  C   ASP B   1      -4.922  -4.362  -3.009  1.00  0.00           C  
ATOM     14  O   ASP B   1      -4.495  -3.371  -3.582  1.00  0.00           O  
ATOM     15  CB  ASP B   1      -2.583  -5.335  -2.926  1.00  0.00           C  
ATOM     16  CG  ASP B   1      -2.042  -4.468  -1.808  1.00  0.00           C  
ATOM     17  OD1 ASP B   1      -2.752  -4.286  -0.791  1.00  0.00           O  
ATOM     18  OD2 ASP B   1      -0.880  -4.003  -1.955  1.00  0.00           O  
TER
ATOM     19  N   GLY C   1       5.361   5.214   1.901  1.00  0.00           N  
ATOM     20  CA  GLY C   1       4.400   6.252   2.188  1.00  0.00           C  
ATOM     21  C   GLY C   1       3.501   6.577   1.006  1.00  0.00           C  
ATOM     22  O   GLY C   1       2.479   7.251   1.188  1.00  0.00           O  
ATOM     23  N   UNK C   2       3.821   6.075  -0.188  1.00  0.00           N  
ATOM     24  CA  UNK C   2       2.965   6.247  -1.351  1.00  0.00           C  
ATOM     25  C   UNK C   2       1.905   5.154  -1.494  1.00  0.00           C  
ATOM     26  O   UNK C   2       0.929   5.369  -2.228  1.00  0.00           O  
ATOM     27  N   SER C   3       2.100   4.015  -0.827  1.00  0.00           N  
ATOM     28  CA  SER C   3       1.156   2.896  -0.909  1.00  0.00           C  
ATOM     29  C   SER C   3      -0.271   3.352  -0.782  1.00  0.00           C  
ATOM     30  O   SER C   3      -0.604   4.170   0.069  1.00  0.00           O  
ATOM     31  CB  SER C   3       1.374   1.891   0.223  1.00  0.00           C  
ATOM     32  OG  SER C   3       2.357   0.946  -0.128  1.00  0.00           O  
TER
END   
\end{verbatim}
\end{tcolorbox}
    \caption{Example motif specification in PDB format (problem 27, 4XOJ).  }
    \label{fig:example_motif}
\end{figure}
}
\clearpage 
\pagebreak


\section{Evaluation example complete results}\label{sec:demo}
In this part, we provide detailed results described in \nameref{section:baseline_section} in the following manner:
\begin{itemize}
    \item{\textbf{\Cref{table:RFdiff_replicates}} summarizes the RFdiffusion baseline results along with stochasticity test across different replicates for both evaluation and scaffold generation.}
    \item{\textbf{\Cref{table:RFdiff_folding_method}} demonstrates the evaluation results against different structure prediction method. }
    \item{\textbf{\Cref{table:reference_RFdiff,table:reference_RFdiff_contingency}} display the evaluation results on reference scaffolds in comparison with RFdiffusion}
\end{itemize}

\begin{table}[H]
\centering
\caption{RFdiffusion benchmark performances across replicates}
\label{table:RFdiff_replicates}
\footnotesize
\resizebox{\linewidth}{!}
{
\begin{threeparttable}
\begin{tabular}{|p{0.3cm}|p{0.8cm}|p{0.2cm}|p{0.9cm}|p{0.9cm}|p{0.8cm}|p{1.6cm}|p{1.5cm}|p{1.7cm}|p{1.6cm}|p{1.5cm}|p{1.7cm}|}
\hline
\multicolumn{3}{|c|}{\textbf{Case}}               & \multicolumn{3}{c|}{\textbf{Example Run}}                            & \multicolumn{3}{c|}{\textbf{Variability from Benchmark}\tnote{*}} & \multicolumn{3}{c|}{\textbf{Variability from Scaffolds}\tnote{*}}  \\ 
\hline
\# & PDB ID & Group & \# Solutions & Novelty & Success Rate (\%) & \# Solutions & Novelty & Success Rate (\%) & \# Solutions & Novelty & Success Rate (\%) \\ 
\hline
1           & 1LDB            & 1                 & 2                         & 0.369            & 2                     & 2.00 ± 0.00           & 0.36 ± 0.00      & 2.20 ± 0.45                       & 0.60 ± 0.89           & 0.16 ± 0.22      & 0.60 ± 0.89                        \\
2           & 1ITU            & 1                 & 2                         & 0.324            & 4                     & 2.20 ± 0.45           & 0.33 ± 0.01      & 4.20 ± 0.45                       & 2.20 ± 0.45           & 0.33 ± 0.01      & 3.81 ± 1.48                        \\
3           & 2CGA            & 1                 & 0                         & 0                & 0                     & 0.00 ± 0.00           & 0.00 ± 0.00      & 0.00 ± 0.00                       & 0.00 ± 0.00           & 0.00 ± 0.00      & 0.00 ± 0.00                        \\
4           & 5WN9            & 1                 & 10                        & 0.336            & 11                    & 9.80 ± 0.45           & 0.34 ± 0.01      & 10.80 ± 0.45                      & 6.60 ± 2.41           & 0.33 ± 0.07      & 6.80 ± 2.77                        \\
5           & 5ZE9            & 1                 & 27                        & 0.355             & 31                    & 27.25 ± 0.96          & 0.36 ± 0.00      & 31.40 ± 0.71 & 26.00 ± 1.41 & 0.36 ± 0.01      & 31.40 ± 0.55                       \\
6           & 6E6R            & 1                 & 44                        & 0.316            & 72                    & 44.00 ± 0.00          & 0.32 ± 0.00      & 72.00 ± 0.00                      & 45.80 ± 2.49          & 0.31 ± 0.00      & 77.70 ± 5.12                       \\
7           & 6E6R            & 1                 & 74                        & 0.388            & 75                    & 73.75 ± 0.50          & 0.40 ± 0.00      & 74.94 ± 0.12                      & 75.25 ± 3.20          & 0.40 ± 0.00      & 76.00 ± 3.37                       \\
8           & 7AD5            & 1                 & 0                         & 0                & 0                     & 0.00 ± 0.00           & 0.00 ± 0.00      & 0.00 ± 0.00                       & 0.00 ± 0.00           & 0.00 ± 0.00      & 0.00 ± 0.00                        \\
9           & 7CG5            & 1                 & 32                        & 0.376            & 74                    & 31.50 ± 0.58          & 0.38 ± 0.00      & 73.50 ± 0.58                      & 31.75 ± 0.50          & 0.37 ± 0.01      & 73.25 ± 0.96                       \\
10          & 7WRK            & 1                 & 0                         & 0                & 0                     & 0.00 ± 0.00           & 0.00 ± 0.00      & 0.00 ± 0.00                       & 0.00 ± 0.00           & 0.00 ± 0.00      & 0.00 ± 0.00                        \\ 
\hline \hline
11          & 3TQB            & 2                 & 55                        & 0.405            & 96                    & 55.00 ± 0.00          & 0.40 ± 0.00      & 96.00 ± 0.00                      & 55.50 ± 0.71          & 0.41 ± 0.01      & 96.00 ± 0.00                       \\
12          & 4JHW            & 2                 & 0                         & 0                & 0                     & 0.00 ± 0.00           & 0.00 ± 0.00      & 0.00 ± 0.00                       & 0.00 ± 0.00           & 0.00 ± 0.00      & 0.00 ± 0.00                        \\
13          & 4JHW            & 2                 & 0                         & 0                & 0                     & 0.00 ± 0.00           & 0.00 ± 0.00      & 0.00 ± 0.00                       & 0.00 ± 0.00           & 0.00 ± 0.00      & 0.00 ± 0.00                        \\
14          & 5IUS            & 2                 & 4                         & 0.367            & 25                    & 2.40 ± 0.89           & 0.38 ± 0.00      & 25.50 ± 1.22                      & 3.20 ± 1.10           & 0.36 ± 0.01      & 28.00 ± 0.71                       \\
15          & 7A8S            & 2                 & 2                         & 0.417            & 12                    & 2.00 ± 0.00           & 0.41 ± 0.00      & 12.02 ± 0.05                      & 2.40 ± 0.89           & 0.42 ± 0.01      & 11.40 ± 1.52                       \\
16          & 7BNY            & 2                 & 2                         & 0.434            & 9                     & 2.00 ± 0.00           & 0.42 ± 0.00      & 9.20 ± 0.45                       & 3.00 ± 1.41           & 0.44 ± 0.02      & 8.50 ± 1.73                        \\
17          & 7DGW            & 2                 & 1                         & 0.138            & 37                    & 1.40 ± 0.55           & 0.14 ± 0.01      & 37.20 ± 0.45                      & 1.20 ± 0.45           & 0.14 ± 0.01      & 36.67 ± 0.47                       \\
18          & 7MQQ            & 2                 & 1                         & 0.455            & 4                     & 1.00 ± 0.00           & 0.45 ± 0.00      & 4.01 ± 0.02                       & 1.20 ± 0.45           & 0.44 ± 0.03      & 4.21 ± 0.44                        \\
19          & 7MQQ            & 2                 & 7                         & 0.426            & 8                     & 7.00 ± 0.00           & 0.41 ± 0.00      & 8.02 ± 0.04                       & 5.75 ± 1.50           & 0.40 ± 0.03      & 7.25 ± 0.96                        \\
20          & 7UWL            & 2                 & 0                         & 0                & 0                     & 0.00 ± 0.00           & 0.00 ± 0.00      & 0.00 ± 0.00                       & 0.00 ± 0.00           & 0.00 ± 0.00      & 0.00 ± 0.00                        \\ 
\hline \hline
21          & 1B73            & 3                 & 0                         & 0                & 0                     & 0.00 ± 0.00           & 0.00 ± 0.00      & 0.00 ± 0.00                       & 0.00 ± 0.00           & 0.00 ± 0.00      & 0.00 ± 0.00                        \\
22          & 1BCF            & 3                 & 1                         & 0.167            & 100                   & 1.00 ± 0.00           & 0.17 ± 0.00      & 100.0 ± 0.00 & 1.00 ± 0.00 & 0.17 ± 0.00 & 100.0 ± 0.00 \\
23          & 1MPY            & 3                 & 0                         & 0                & 0                     & 0.00 ± 0.00           & 0.00 ± 0.00      & 0.00 ± 0.00                       & 0.00 ± 0.00           & 0.00 ± 0.00      & 0.00 ± 0.00                        \\
24          & 1QY3            & 3                 & 0                         & 0                & 0                     & 0.00 ± 0.00           & 0.00 ± 0.00      & 0.00 ± 0.00                       & 0.00 ± 0.00           & 0.00 ± 0.00      & 0.00 ± 0.00                        \\
25          & 2RKX            & 3                 & 0                         & 0                & 0                     & 0.00 ± 0.00           & 0.00 ± 0.00      & 0.00 ± 0.00                       & 0.00 ± 0.00           & 0.00 ± 0.00      & 0.00 ± 0.00                        \\
26          & 3B5V            & 3                 & 0                         & 0                & 0                     & 0.00 ± 0.00           & 0.00 ± 0.00      & 0.00 ± 0.00                       & 0.00 ± 0.00           & 0.00 ± 0.00      & 0.00 ± 0.00                        \\
27          & 4XOJ            & 3                 & 0                         & 0                & 0                     & 0.00 ± 0.00           & 0.00 ± 0.00      & 0.00 ± 0.00                       & 0.00 ± 0.00           & 0.00 ± 0.00      & 0.00 ± 0.00                        \\
28          & 5YUI            & 3                 & 3                         & 0.377            & 7                     & 3.00 ± 0.00           & 0.43 ± 0.00      & 7.41 ± 0.54                       & 3.40 ± 0.55           & 0.41 ± 0.02      & 6.80 ± 0.84                        \\
29          & 6CPA            & 3                 & 0                         & 0                & 0                     & 0.00 ± 0.00           & 0.00 ± 0.00      & 0.00 ± 0.00                       & 0.00 ± 0.00           & 0.00 ± 0.00      & 0.00 ± 0.00                        \\
30          & 7UWL            & 3                 & 0                         & 0                & 0                     & 0.00 ± 0.00           & 0.00 ± 0.00      & 0.00 ± 0.00                       & 0.00 ± 0.00           & 0.00 ± 0.00      & 0.00 ± 0.00                        \\ 
\hline
\multicolumn{3}{|c|}{\textbf{MotifBench Score}}   & 28.05 & & & 28.37 ± 0.39 & & & 27.80 ± 0.47 & & \\
\hline
\end{tabular}
\begin{tablenotes}
    \footnotesize
    \item[*] Values denoted as ``Mean ± Standard deviation''.
\end{tablenotes}
\end{threeparttable}
}
\end{table}

\begin{table}[p]
\centering
\caption{RFdiffusion benchmark metrics with ESMFold and AlphaFold2}
\label{table:RFdiff_folding_method}
\resizebox{\linewidth}{!}{
\begin{tabular}{|l|l|l|lll|lll|} 
\hline
\multicolumn{3}{|c|}{\textbf{Case}}               & \multicolumn{3}{c|}{\textbf{ESMFold}}        & \multicolumn{3}{c|}{\textbf{AlphaFold2 (with no MSA input)}}                \\ 
\hline
\# & PDB ID & Group & \# Solutions & Novelty & Success Rate (\%) & \# Solutions & Novelty & Success Rate (\%) \\ 
\hline
1           & 1LDB        & 1                 & 2                         & 0.369            & 2                     & 0                     & 0                & 0                                 \\
2           & 1ITU        & 1                 & 2                         & 0.324            & 4                     & 1                     & 0.367            & 1                                 \\
3           & 2CGA        & 1                 & 0                         & 0                & 0                     & 0                     & 0                & 0                                 \\
4           & 5WN9        & 1                 & 10                        & 0.336            & 11                    & 21                    & 0.297            & 24                                \\
5           & 5ZE9        & 1                 & 27                        & 0.355            & 31                    & 0                     & 0                & 0                                 \\
6           & 6E6R        & 1                 & 44                        & 0.316            & 72                    & 47                    & 0.306            & 76                                \\
7           & 6E6R        & 1                 & 74                        & 0.388            & 75                    & 67                    & 0.395            & 69                                \\
8           & 7AD5        & 1                 & 0                         & 0                & 0                     & 0                     & 0                & 0                                 \\
9           & 7CG5        & 1                 & 32                        & 0.376            & 74                    & 21                    & 0.372            & 55                                \\
10          & 7WRK        & 1                 & 0                         & 0                & 0                     & 0                     & 0                & 0                                 \\ 
\hline \hline
11          & 3TQB        & 2                 & 55                        & 0.405            & 96                    & 56                    & 0.406            & 95                                \\
12          & 4JHW        & 2                 & 0                         & 0                & 0                     & 0                     & 0                & 0                                 \\
13          & 4JHW        & 2                 & 0                         & 0                & 0                     & 0                     & 0                & 0                                 \\
14          & 5IUS        & 2                 & 4                         & 0.367            & 25                    & 4                     & 0.367            & 29                                \\
15          & 7A8S        & 2                 & 2                         & 0.417            & 12                    & 0                     & 0                & 0                                 \\
16          & 7BNY        & 2                 & 2                         & 0.434            & 9                     & 3                     & 0.456            & 6                                 \\
17          & 7DGW        & 2                 & 1                         & 0.138            & 37                    & 0                     & 0                & 0                                 \\
18          & 7MQQ        & 2                 & 1                         & 0.455            & 4                     & 1                     & 0.442            & 2                                 \\
19          & 7MQQ        & 2                 & 7                         & 0.426            & 8                     & 2                     & 0.460            & 2                                 \\
20          & 7UWL        & 2                 & 0                         & 0                & 0                     & 0                     & 0                & 0                                 \\ 
\hline \hline
21          & 1B73        & 3                 & 0                         & 0                & 0                     & 0                     & 0                & 0                                 \\
22          & 1BCF        & 3                 & 1                         & 0.167            & 100                   & 1                     & 0.167            & 100                               \\
23          & 1MPY        & 3                 & 0                         & 0                & 0                     & 0                     & 0                & 0                                 \\
24          & 1QY3        & 3                 & 0                         & 0                & 0                     & 0                     & 0                & 0                                 \\
25          & 2RKX        & 3                 & 0                         & 0                & 0                     & 0                     & 0                & 0                                 \\
26          & 3B5V        & 3                 & 0                         & 0                & 0                     & 0                     & 0                & 0                                 \\
27          & 4XOJ        & 3                 & 0                         & 0                & 0                     & 0                     & 0                & 0                                 \\
28          & 5YUI        & 3                 & 3                         & 0.377            & 7                     & 4                     & 0.379            & 8                                 \\
29          & 6CPA        & 3                 & 0                         & 0                & 0                     & 0                     & 0                & 0                                 \\
30          & 7UWL        & 3                 & 0                         & 0                & 0                     & 0                     & 0                & 0                                 \\ 
\hline
\end{tabular}
}
\end{table}

\begin{table}
\centering
\caption{Evaluation of Reference Scaffolds Compared with RFdiffusion}
\label{table:reference_RFdiff}
\resizebox{\linewidth}{!}{
\begin{tabular}{|l|l|l|l|l|l|p{2.0cm}|p{2.0cm}|}
\hline
\# & PDB ID & Group   & RFdiffusion \# solutions & Reference structure passes & RFdiffusion solves & Minimum motif-RMSD of reference scaffold (\AA) & Minimum sc-RMSD of reference scaffold (\AA) \\
\hline
1  & 1LDB   & 1       & 2                         & \Checkmark                 & \Checkmark  & 0.478 & 1.242        \\
2  & 1ITU   & 1       & 2                         & \Checkmark                 & \Checkmark  & 0.21 & 1.226      \\
3  & 2CGA   & 1       & 0                         & \XSolidBrush               & \XSolidBrush  & \textbf{2.05} & 1.102               \\
4  & 5WN9   & 1       & 10                        & \Checkmark                 & \Checkmark  & 0.434 & 0.323                   \\
5  & 5ZE9   & 1       & 27                        & \Checkmark                 & \Checkmark  & 0.257 & 1.01          \\
6  & 6E6R   & 1       & 44                        & \Checkmark                 & \Checkmark  & 0.218 & 0.325          \\
7  & 6E6R   & 1       & 74                        & \Checkmark                 & \Checkmark  & 0.218 & 0.325          \\
8  & 7AD5   & 1       & 0                         & \XSolidBrush               & \XSolidBrush  & \textbf{2.48} & 2.156        \\
9  & 7CG5   & 1       & 32                        & \Checkmark                 & \Checkmark  & 0.156 & 1.066          \\
10 & 7WRK   & 1       & 0                         & \Checkmark                 & \XSolidBrush  & 0.219 & 0.68         \\
\hline \hline
11 & 3TQB   & 2       & 55                        & \Checkmark                 & \Checkmark  & 0.607 & 0.523          \\
12 & 4JHW   & 2       & 0                         & \XSolidBrush               & \XSolidBrush  & \textbf{2.059} & \textbf{16.413}       \\
13 & 4JHW   & 2       & 0                         & \XSolidBrush               & \XSolidBrush  & \textbf{2.059} &\textbf{16.413}        \\
14 & 5IUS   & 2       & 4                         & \Checkmark                 & \Checkmark  & 0.868 & 0.884          \\
15 & 7A8S   & 2       & 2                         & \Checkmark                 & \Checkmark  & 0.344 & 0.781          \\
16 & 7BNY   & 2       & 2                         & \Checkmark                 & \Checkmark  & 0.818 & 1.688          \\
17 & 7DGW   & 2       & 1                         & \Checkmark                 & \Checkmark  & 0.478 & 0.451          \\
18 & 7MQQ   & 2       & 1                         & \XSolidBrush               & \Checkmark  & \textbf{1.213} & 1.649           \\
19 & 7MQQ   & 2       & 7                         & \XSolidBrush               & \Checkmark  & \textbf{1.213} & 1.649           \\
20 & 7UWL   & 2       & 0                         & \XSolidBrush               & \XSolidBrush  &\textbf{1.657} & \textbf{3.627}        \\
\hline \hline
21 & 1B73   & 3       & 0                         & \XSolidBrush               & \XSolidBrush  & \textbf{1.313} & 1.286        \\
22 & 1BCF   & 3       & 1                         & \Checkmark                 & \Checkmark  & 0.289 & 0.803          \\
23 & 1MPY   & 3       & 0                         & \Checkmark                 & \XSolidBrush  & 0.28 & 0.975         \\
24 & 1QY3   & 3       & 0                         & \XSolidBrush               & \XSolidBrush  & \textbf{1.498} & \textbf{2.291}        \\
25 & 2RKX   & 3       & 0                         & \Checkmark                 & \XSolidBrush  & 0.276 & 0.475         \\
26 & 3B5V   & 3       & 0                         & \Checkmark                 & \XSolidBrush  & 0.348 & 0.661         \\
27 & 4XOJ   & 3       & 0                         & \Checkmark                 & \XSolidBrush  & 0.167 & 0.424         \\
28 & 5YUI   & 3       & 3                         & \Checkmark                 & \Checkmark  & 0.462 & 0.591          \\
29 & 6CPA   & 3       & 0                         & \Checkmark                 & \XSolidBrush  & 0.25 &  0.575         \\
30 & 7UWL   & 3       & 0                         & \XSolidBrush               & \XSolidBrush  & \textbf{1.717} & \textbf{3.627}        \\
\hline
\multicolumn{1}{|l}{} & \multicolumn{1}{l}{} & \multicolumn{1}{l}{} & \multicolumn{1}{l|}{}    & \multicolumn{1}{c}{}       & \multicolumn{1}{l}{} & \multicolumn{1}{c}{}                         &                                            \\ 
\hline
\multicolumn{3}{|l}{\textbf{\# Reference structure passes}}                  &                           & \multicolumn{4}{c|}{20}                          \\ 
\hline
\multicolumn{3}{|l}{\textbf{\# RFdiffusion solves}}                          &                           & \multicolumn{4}{c|}{16}                          \\
\hline
\end{tabular}
}
\end{table}

\begin{table}
\centering
\caption{Contingency Results between Reference Structures and RFdiffusion}
\label{table:reference_RFdiff_contingency}
\begin{tabular}{|c|c|c|} 
\hline
\diagbox{\textbf{Reference Structure}}{\textbf{RFdiffusion}} & Pass & Fail  \\ 
\hline
Pass                                                         & 14   & 6     \\ 
\hline
Fail                                                         & 2    & 8     \\
\hline
\end{tabular}
\end{table}

\end{document}